\documentclass{article}
\usepackage{xr}

\usepackage[preprint,nonatbib]{neurips_2020}
\usepackage[utf8]{inputenc} 
\usepackage[T1]{fontenc}    
\usepackage{hyperref}       
\usepackage{url}            
\usepackage{booktabs}       
\usepackage{amsfonts}       
\usepackage{nicefrac}       
\usepackage{microtype}      

\usepackage{hyperref}

\usepackage{amssymb}

\newcommand{\intCP}[2]{\mathcal{T}{(#1;#2)}}
\newcommand{\prob}[1]{\mathbb{P}{(#1)}}
\newcommand{\intRA}[1]{\mathcal{I}{(#1)}}
\newcommand{\mcal}[1]{\mathcal{#1}}
\newcommand{\Z}{\mcal Z}
\newcommand{\E}{\mathbb{E}}
\newcommand{\I}{\mcal I}

\usepackage{mathtools}
\DeclarePairedDelimiter\ceil{\lceil}{\rceil}

\usepackage{amsmath}
\usepackage{amssymb}
\usepackage{mathtools}
\usepackage{amsthm}
\usepackage{multirow}
\usepackage{breqn}
\usepackage{caption}
\usepackage{subcaption}
\usepackage{tikz}
\usetikzlibrary{arrows, arrows.meta}
\usepackage{bbm}
\usepackage[capitalize,noabbrev]{cleveref}
\theoremstyle{plain}
\newtheorem{theorem}{Theorem}[section]

\theoremstyle{definition}

\theoremstyle{remark}
\newtheorem{remark}[theorem]{Remark}

\usepackage[textsize=tiny]{todonotes}

\title{Distribution-free risk assessment of regression-based machine learning algorithms}

\author{%
  Sukrita Singh \\
  University of Oxford\\
  \texttt{sukrita.singh@sbs.ox.ac.uk} \\
  \And
  Neeraj Sarna \\
  Munich Re \\
  \texttt{nsarna@munichre.com} \\
  \AND
  Yuanyuan Li \\
  Munich Re \\
  \texttt{yli@munichre.com} \\
  \And
  Yang Lin \\
  Munich Re \\
  \texttt{ylin@munichre.com} \\
  \And
  Agni Orfanoudaki \\
  University of Oxford\\
  \texttt{agni.orfanoudaki@sbs.ox.ac.uk} \\
    \And
  Michael Berger \\
  Munich Re\\
  \texttt{mberger@munichre.com} \\
}

\begin{document}

\maketitle

\begin{abstract}
Machine learning algorithms have grown in sophistication over the years and are increasingly deployed for real-life applications. However, when using machine learning techniques in practical settings, particularly in high-risk applications such as medicine and engineering, obtaining the failure probability of the predictive model is critical. We refer to this problem as the risk-assessment task. We focus on regression algorithms and the risk-assessment task of computing the probability of the true label lying inside an interval defined around the model's prediction. We solve the risk-assessment problem using the conformal prediction approach, which provides prediction intervals that are guaranteed to contain the true label with a given probability. Using this coverage property, we prove that our approximated failure probability is conservative in the sense that it is not lower than the true failure probability of the ML algorithm. We conduct extensive experiments to empirically study the accuracy of the proposed method for problems with and without covariate shift. Our analysis focuses on different modeling regimes, dataset sizes, and conformal prediction methodologies.
\end{abstract}

\section{Introduction}
\label{intro}

In safety-critical applications, it is crucial that the ML modelling errors stay within certain limits. One is then interested in computing the probability of the errors being larger than these limits. Consider, for instance, a ML model that predicts the health of a battery \cite{battery_1,battery_2}. A battery owner using such a model would then be interested in the probability of the modelling error being larger than a threshold. This would help the owner assess the extent to which a battery's health might be jeoperdized, in case the model under-performs. Similar scenarios could also arise in safety-critical medical applications where a model predicting medicine dosage cannot be more than $10\%$ inaccurate \cite{radiationML}. Solving this critical challenge of ML risk evaluation can have drastic implications on the degree of adoption of such algorithms in practice. We refer to this challenge as the risk-assessment problem.
\subsection{Risk-assessment: problem formulation}
We formalize the above problem mathematically. Consider a given pre-defined interval $\intRA{X} = \left[a_-(X),a_+(X)\right]$, where $a_\pm(X)\in\mathbb {R}$ and $a_-(X) < a_+(X)$. The risk-assessment problem seeks the miscoverage rate $\alpha_{\mathcal{I}}$ that approximates $\prob{Y\not\in \intRA{X}}$, where $X\in\mathbb{R}^d$ and $Y\in\mathbb{R}$ denote the input and output, respectively. Equivalently,
\begin{gather}
\text{\textit{Risk assessment task:}}\quad\quad\text{Given $\I(X)$ find $\alpha_\I$ : } \prob{Y\in \intRA{X}}\geq 1-\alpha_\I, \label{RA problem}
\end{gather}
We require two properties from $\alpha_{\I}$. Firstly, we require accurate risk-assessment. For a reliable risk-assessment, the coverage $1-\alpha_{\mathcal{I}}$ should be an accurate approximation of $\prob{Y\in \intRA{X}}$. Secondly, we require conservative risk-assessment. The probability $\prob{Y\in \intRA{X}}$ should be bounded from below by $1-c\alpha_{\mathcal{I}}$, which ensures that we do not over-estimate this probability \footnote{The value of $c$ would depend on the prediction interval approach---details discussed later.}.
Note that conservative risk-assessment does not necessarily provide an accurate risk-assessment or vice-versa. For example, $\alpha_\I = 1$ -- i.e. a $100\%$ failure rate -- results in a conservative risk-assessment for any ML model but for most ML models, provides inaccurate risk-assessment.  

We seek a distribution-free solution to the risk-assessment problem. The following argument motivates our choice. For classical probabilistic models with assumptions about the data distribution, achieving the risk assessment task is straightforward. For example, we consider a linear regression model fitted by least-square-method. If we assume the prediction errors $\hat{\epsilon}=Y-\hat{Y}$ given by the predictor $\hat{Y} \in \mathbb{R}$ follows a normal distribution $P_{\hat{\epsilon}}$ and are independent with $X$, we get immediately $\alpha_I=\int_{Y \not\in I(X)}dP_{\hat{\epsilon}}$. However, in many cases in practice such assumptions generally do not hold. For solving the risk-assessment task across many use cases, we do not make distributional assumptions, i.e. we work in a distribution-free setting. To this end, we exploit distribution-free prediction interval generation.

\subsection{Prediction interval generation}\label{sec: PI generation}
In prediction interval generation, we are given a miscoverage level $\alpha$ and we compute a corresponding prediction interval $\intCP{X}{\alpha}$. 
We leverage this to solve our risk-assessment problem. The idea is to generate nested prediction intervals with varying miscoverage rates and select the largest prediction interval that fits inside the given interval $\I(X)$. The miscoverage of this largest interval then provides a solution to the risk-assessment problem---see \autoref{sec: previous work} for details. Furthermore, as discussed in \autoref{sec: previous work}, this nested property helps us recover a conservative risk-assessment.

The problem of prediction interval generation can be precisely posed as 
\begin{gather} 
\text{\textit{Prediction interval computation:}}\quad\quad\text{Given $\alpha$ find $\intCP{X}{\alpha}$ : } \prob{Y\in \intCP{X}{\alpha}} \geq 1-c\alpha, \label{PI problem}
\end{gather}
where $X\in\mathbb{R}^d$ and $Y\in\mathbb{R}$ represent the input and label, respectively, and the value of $c$ depends upon the method under consideration. Note that the probability in the above expression is not conditional over a particular test point but instead, marginal---it captures the randomness over the entire test set. Indeed, in general, one could prove that conditional coverage cannot be achieved \cite{conditionalCP}. This point will be crucial when we later compare our method to previous works. 

Our need for distribution-free risk-assessment, motivates choosing conformal prediction(CP) as our prediction interval generation technique---\autoref{sec:background} provides a brief overview to CP. CP relies on a (so-called) conformal score, which encapsulates the heuristic uncertainty of a model---an example being the residual for a regression algorithm.  The prediction interval has provable coverage properties, i.e. it satisfies the lower bound on the probability given in \autoref{PI problem} \cite{conformal_pred_intro, conformal_pred}. The quality of the prediction interval is largely determined by the choice of the score function \cite{cp_res_score, cqr, score_function_1,score_function_2,score_function_3}. The CP technique is model agnostic and can be applied to both regression \cite{split_cp_1, split_adapt} and classification problems \cite{classification_avg_size,cp_classficiation_image}. Although the initial work was limited to exchangeable datasets \cite{conformal_pred_intro, old_tutorial}, the technique has been recently extended to non-exchangeable data \cite{CP_covariate, beyond_exchangebility}. We briefly review the CP technique in \autoref{sec:background}.

\subsection{Previous work on distribution-free risk assessment}
In a distribution-free setting, to the best of our knowledge, only the authors in \cite{JAWS} solve the risk assessment problem using CP. They refer to their CP algorithm as JAW, which provides the marginal coverage property described in \autoref{PI problem}. \autoref{sec: previous work} provides the details. The main shortcomings are summarised here. Firstly, the JAW-based risk assessment approach -- although claimed to be conservative in the sense defined in \autoref{intro} -- is not conservative theoretically. Secondly, the JAW-based approach disregards the randomness in the solution to the risk-assessment task and thereby, compared to our solution, provides a crude high-variance solution to the risk assessment problem -- \autoref{sec: theory} elaborates further. Lastly, authors in \cite{JAWS} do not numerically analyse the accuracy of their solution to the risk-assessment problem. An AUC-type approach was explored and as yet, it is unclear how this relates to the accuracy or to how conservative the risk-assessment is.

\subsection{Contributions}
Our contributions, which cater to the problems encountered by the JAW-based approach, are summarized as follows. Firstly, we formalize the risk-assessment task and we propose a general-purpose framework to solve the risk-assessment task for regression problems that leverages CP techniques. Using the coverage property of the CP technique, we prove that our risk-assessment is conservative in the sense described in \autoref{intro}. We also observe experimentally that our risk-assessment is conservative. Secondly, we identify the randomness in the solution to the risk-assessment task and capture it accurately via a hold-out set. Our hold-out set does not require label information and thus, could also leverage  generative-ML techniques \cite{Goodfellow-book}. Lastly, to assess the accuracy of our risk-assessment algorithm, we propose a comprehensive set of computational experiments on problems with and without covariate shifts. We assess the influence of model-type, data size, and CP technique on the proposed algorithm.

\section{Background}
\label{sec:background}
As mentioned in \autoref{intro}, we choose CP as our prediction interval generation technique. This section briefly summarises CP in the covariate shift setting. This setting is more general and more practically relevant than the i.i.d setting (or the weaker exchangeability setting) considered earlier by the CP literature. We start with introducing covariate shift.

\subsection{Covariate shift}\label{sec: cov shift}
The covariate shift problem has gained attention within the field of uncertainty quantification in recent studies \cite{CP_covariate, beyond_exchangebility, covariateML, covariate_shift_1}. This topic has also been highlighted in various ML applications, specifically in the context of health-related applications\cite{data_shift_1,data_shift_2,data_shift_3}. Under covariate shift, the distribution for $Y|X$ remains the same under training and testing. However, the distribution for $X$ could change. Let $P_{Y|X}$ and $P_X$ represent the distribution for $Y|X$ and $X$, respectively, under training. Under testing, the distribution for $X$ changes to $\tilde{P}_X$. The training and the testing data points are sampled from their respective distributions independently. This data setting clearly violates the important data exchangeability assumption CP based on. To adapt the standard CP techniques for covariate shift, Tibshirani et.al \cite{CP_covariate} proposed the concept of \textit{weight exchangeability}, and proved that if $\tilde P_X$ is absolutely continuous with respect to $P_X$, the data under the covariate shift are weighted exchangeble. By utilizing a weight function, i.e., the likelihood ratio of testing covariate distribution over the training one $ w(x) = \mathrm{d}\tilde{P}_X(x)/\mathrm{d}P_X(x) \label{weight fun}$, the weighted CP intervals can provide a valid coverage guarantee under covariate shift \cite{CP_covariate}. We present more details in the next section.

\subsection{Conformal Prediction (CP)}\label{sec: cp}

We restrict ourselves to the weighted split-CP \cite{CP_covariate} and JAW \cite{JAWS} approaches which are applicable to the co-variate shift setting. We first discuss the split-CP technique. Consider a score-function $S: \mathbb{R}^d \times \mathbb{R} \rightarrow \mathbb{R}$, and
$S(x,y) = |y - \mu(x)|$,
where $\mu(x) : \mathbb{R}^d \rightarrow \mathbb{R}$ represents our model's approximation at point $x$. Let $\mcal Z = \left\{Z_i = (X_i,Y_i)\right\}_{i=1,\dots,n}$ represent a set of calibration points that are independent from the set used to train $\mu$. For an input point $x$, the weighted split-CP prediction interval is then defined as
$
    \intCP{x}{\alpha} = \mu(x) \pm Q^+_{\alpha} \{p_i^w(x)\delta_{S(X_i,Y_i)}\}.
$
For notational simplicity, we suppress the dependence of $\mcal T$ on $\mcal Z$. The weights $p_i^w(x)$ and $p_{n+1}^w(x)$ are defined as
\begin{gather}
    p_i^w(x) = \frac{w(X_i)}{\sum_{j=1}^nw(X_j) + w(x)}, i=1, \cdots, n, \quad p_{n+1}^w(x) = \frac{w(x)}{\sum_{j=1}^nw(X_j) + w(x)},\label{weights}
\end{gather}
where $w(X)$ is defined in \autoref{sec: cov shift}. The term $Q^+_{\alpha}\{v_i\}$ is the $\ceil{(1-\alpha)(n+1)}$-th smallest value of $v_1,v_2,\dots,v_n$, and $Q$ is the quantile function under the empirical distribution of the values $v_1, \cdots, v_n$. The $\delta_v$ represents the point mass distribution at point $v$. The idea behind introducing the weight functions is weight exchangeability, allowing one to use exchangeability-based tools from standard CP\cite{CP_covariate}.

The JAW approach collects samples of the score function differently. It considers a leave-one-out approach on the training data set. Furthermore, let $\mu_{-i}$ represent a model trained on all training points other than the i-th one. Samples of the score function are then defined as $S(X_i,Y_i) = |Y_i - \mu_{-i}(X_i)|$. Furthermore, in the formulae for the prediction interval given above, JAW replaces $\mu$ by $\mu_{-i}$. The prediction interval for JAW then reads
    $\intCP{x}{\alpha} = \left[Q^-_{\alpha}\left\{p_i^w(x)\delta_{\mu_{-i}(x) - S(X_i,Y_i)}\right\},Q^+_{\alpha}\left\{p_i^w(x)\delta_{\mu_{-i}(x) + S(X_i,Y_i)}\right\}\right]$
where $Q^-_{\alpha}\{v_i\}$ is the $\ceil{\alpha(n+1)}$-th smallest value of $v_1,v_2,\dots,v_n$. For simplicity, we collectively express the above two prediction intervals as
\begin{gather}
   \intCP{x}{\alpha} = \left[Q^-_{\alpha}\left\{p_i^w(x)\delta_{V^{-}_{i}(X)}\right\},Q^+_{\alpha}\left\{p_i^w(x)\delta_{V^{+}_{i}(X)}\right\}\right], \label{interval gen}
\end{gather}
where $V^{+}_{i}(X)$ and $V^{-}_{i}(X)$ are the upper and lower point masses, respectively, and read
\begin{gather}
  V^{\pm}_{i}(X):= \mu_{\square}
  (X) \pm S(X_i,Y_i). \label{def Vpm}
\end{gather}
The quantity $(\square)$ is a placeholder which could either be empty or $-i$ for split-CP and JAW, respectively. Recall that $-i$ represents that the $i$-th training point was dropped while training $\mu_{-i}$.

The following properties are noteworthy. Firstly, for exchangeable datasets, substituting $w=1$, the weighted split-CP and the JAW intervals reduce to prediction intervals that correspond to the standard split-CP and the Jackknife+ intervals, respectively. Secondly, one can show that the above prediction intervals are nested in the sense that
\begin{gather}
    \intCP{X}{\alpha_1} \subseteq \intCP{X}{\alpha_2},\quad \forall \alpha_1\geq \alpha_2. \label{eq: nested}
\end{gather}
The above property would be helpful later during our risk-assessment formulation.
Secondly, both the weighted split-CP and JAW have the coverage property, which, for later convenience, we summarize below.

\begin{theorem} \label{thrm: CP}
Under the assumptions: a) data under co-variate shift; and b)$\tilde P_X$ is absolutely continuous with respect to $P_X$, the interval predictions resulting from weighted split-CP and JAW satisfy
$
    \prob{Y\in \intCP{X}{\alpha}} \geq 1-c\alpha,
$
where $c$ equals $1$ and $2$ for split-CP and JAW, respectively.
\end{theorem}
\textit{Proof}: 
See \cite{CP_covariate,JAWS}.\hfill $\square$

\section{Proposed approach: theoretical properties and algorithm}
\subsection{Nested prediction interval generation}\label{sec: previous work}
We discuss how nested prediction intervals help us solve the risk-assessment problem. We generate nested prediction intervals that are contained inside the pre-defined interval $\intRA{X}$ and use the coverage of these prediction intervals to solve the risk-assessment problem. Building upon the idea from \cite{JAWS}, we express the nested interval generation as follows.

For any test input $X$ and a calibration set $\mcal Z$, we seek a miscoverage $\alpha(X,\Z)$ such that the prediction interval (given in \autoref{interval gen}), at the test input, is contained inside the pre-defined interval $\intRA{X} = \left[a_-(X),a_+(X)\right]$. Equivalently, 
 \begin{equation}
        \alpha\left(X,\Z\right): = 
        \min_{\alpha^{'}}\{\alpha^{'}: \intCP{X}{\alpha'} \subseteq \intRA{X}\}.\label{alphaXZ}
    \end{equation}
where the prediction interval $\intCP{x}{\alpha'}$ is computed using CP, which uses the calibration set $\mcal Z$. Two points motivate the above definition of $\alpha(X,\Z)$. Firstly, since $\intCP{X}{\alpha(X,\Z)}$ is included in $\intRA{X}$, we conclude that 
\begin{gather}
    \prob{Y\in \intRA{X}|\mcal \Z = \mcal Z_0, X=X^*} \geq \prob{Y\in \intCP{X}{\alpha(X,\Z)}|\mcal \Z = \mcal Z_0, X=X^*}. \label{condition covg} 
\end{gather}
where $\mcal Z_0$ is a realisation of the calibration set, and the randomness is over $Y|X$.
The above property subsequently provides conservative risk-assessment---see \autoref{sec: main theorem}. Secondly, the minimum over $\alpha'$ (owing to \autoref{eq: nested}) ensures the largest interval $\intCP{X}{\alpha'}$ contained inside $\I(X)$. This ensures the optimality of risk-assessment. Otherwise, one could choose a huge $\alpha'$, resulting in a small $\intCP{X}{\alpha'}$ (that would be included in $\I(X)$) but in a grossly-overconservative and inaccurate failure probability $\mathbb{P}(Y\in \I(X))$.  

Owing to the explicit form of the CP-generated prediction interval $\intCP{X}{\alpha}$ given in \autoref{interval gen} and the nested property in \autoref{eq: nested}, solving for $\alpha(X,\Z)$ in \autoref{alphaXZ} involves a simple summing of the masses of those point-masses ($p_i^w(x)\delta_{V^{-}_{i}(X)}$ given in \autoref{interval gen}) whose locations lie inside $\intRA{X}$. This is expressed as
\begin{gather}
\alpha\left(X,\Z\right)=\max(\alpha^-(X),\alpha^+(X)), \text{where} \nonumber\\
 \ \alpha^-(X)=\sum_{i=1}^n p_i^w(X) \mathbbm{1}\{V_{i}^-(X) \leq a_-^*(X)\}, \alpha^+(X)=\sum_{i=1}^n p_i^w(X) \mathbbm{1}\{ a_+^*(X) \leq V_{i}^+(X)\},\label{eq:alpha_XZ_sol}
\end{gather}

where $a_-^*(X)$ is the smallest $V_i^-(X)$ that is greater than $a_-(X)$, and $a_+^*(X)$ is the largest $V_i^+(X)$ that is smaller than $a_+(X)$ for $i=1,\cdots,n$. 

\begin{remark}[Revisiting the bound in \cite{JAWS}]
Authors in \cite{JAWS} erroneously disregard the randomness in the miscoverage $\alpha(X,\Z)$ and the conditionality of the probabilities in \ref{condition covg}. They bound $\prob{Y\in \intCP{X}{\alpha(X,\Z)}|\mcal Z = \mcal Z_0, X=X^*}$ from below by $1-c\alpha(X,\Z)$. As already explained in \autoref{sec: PI generation}, this bound is erroneous because the CP techniques used in \cite{conformal_pred_intro} could only provide marginal coverage. Indeed, in a general setting, conditional coverage could never be achieved \cite{conditionalCP}.

\end{remark}
\subsection{Solution to risk-assessment} \label{sec: theory}
Following the previous discussion, to achieve a conservative risk-assessment, we wish to bound the conditional probabilities in \eqref{condition covg} from below. Since CP provides us marginal coverage, we marginalise these conditional probabilities with respect to the calibration set $\Z$ and the input $X$. This provides
\begin{gather}
    \prob{Y\in \intRA{X}} \geq \prob{Y\in \intCP{X}{\alpha_{\I}}} \geq 1-c\alpha_{\I}, \label{bound alphaI} 
\end{gather}
where 
\begin{gather}
    \alpha_{\I} := \E_{\Z,X}[\alpha(X,\Z)]. \label{def alphaI}  
\end{gather}

The last inequality in the above expression, follows from the coverage property of CP (\autoref{thrm: CP}). 

We consider an unbiased approximation to $\alpha_\I$. This lets us replace $\alpha_\I$ in the above inequality by the average of its approximation, which results in a conservative risk-assessment. For our unbiased estimator, we consider a hold-out set $\Z_0^\alpha:=\{X_{i_{\alpha}}\}_{\{i_\alpha=1,\dots,m\}}$, which is independent from the training and the calibration set $\mcal Z$. We note that this set only represents the testing input distribution and does not require label information.

Using the above, we approximate $\alpha_\I$---and thus $\prob{Y\not\in \intRA{X}}$---via
\begin{gather}
    \alpha_\I \approx \alpha^m_{\I} := \frac{1}{m}\sum_{X\in \Z_0^\alpha}\alpha(X,\mcal Z). \label{def alphaIm}
\end{gather}
\autoref{alg:ICP} summarises the computation of $\alpha_\I^m$. We name this algorithm InvCP (Inverse Conformal Prediction), as it applies the inverse of CP to compute the coverage level instead of a prediction interval. We also discuss the specific cases of this algorithm, see the details in the Supplement S.1.1.

By taking an expectation on both sides of \autoref{def alphaIm} and applying the definition in \autoref{def alphaI}, we can prove that $\alpha_\I^m$ is an unbiased estimator for $\alpha_\I$. Replacing $\alpha_\I$ by the expected value of its estimation in \autoref{bound alphaI}, we find that
\begin{gather} \label{probability_alpha_I_m}
    \prob{Y\in \intRA{X}} \geq 1-c\E_{\Z,X}\left[\alpha_{\I}^m\right].  
\end{gather}
Furthermore, from the law of large numbers, as $m\rightarrow\infty$ and $\forall \Z$, we find $\alpha_\I^m\overset{\mathbb{P}}{\rightarrow} \E_X[\alpha(X,\Z)]$.
We collect our findings in the result below.

\begin{theorem}\label{sec: main theorem}
  Assume that the data has co-variate shift in the sense of \autoref{sec: cov shift}. Furthermore, assume that the prediction interval in the definition for $\alpha(X,\Z)$ (given in \autoref{alphaXZ}) has the coverage property as given in \autoref{thrm: CP}. $\alpha_\I^m$ is an unbiased estimator for $\alpha_I$. Furthermore, the following lower-bound for the probability $\prob{Y\in\intRA{X}}$ 
     \begin{gather}
         \prob{Y\in\intRA{X}} \geq 1-c\alpha_\I =1-c\E_{\Z,X}\left[\alpha_\I^m\right], 
     \end{gather}
     where $c$ equals $1$ and $2$ for split-CP and JAW, respectively. As $m\rightarrow \infty$ and for all $\Z$, the estimator $\alpha_\I^m$ converges, in probability, to the expected value $\E_X[\alpha(X,\Z)]$.
\end{theorem}
\textit{Proof}: 
The derivation of the equations \ref{condition covg}-\ref{probability_alpha_I_m} completes the proof.\hfill$\square$
\vspace{0.1cm}
\begin{remark}[Risk-assessment bound for JAW]
Since we estimate $\mathbb{P}(Y\not \in \I(X))$ using $\alpha_\I^m$, ideally, we expect the bound $\mathbb{P}(Y\not \in \I(X)) \leq \mathbb {E}[\alpha_\I^m]$. As \autoref{sec: main theorem} dictates, this is true for split-CP. However, for JAW, we get the conservative bound $\mathbb{P}(Y\not \in \I(X)) \leq 2\mathbb {E}[\alpha_\I^m]$. This is an artifact of the coverage property of JAW (and also Jackknife+), which reads $\mathbb{P}(Y\in\intCP{X}{\alpha})\geq 1-2\alpha$; see \autoref{thrm: CP}. Nonetheless, in experiments, also for JAW, one observes $c\approx 1$ \cite{JAWS}. We make the same observation for our risk-assessment in \autoref{sec: experiment}.
\end{remark}
\vspace{0.1cm}
\begin{remark}[Connection to previously proposed bounds]\label{remark: previous work}
In light of the above theorem, we comment on the theoretical comparison between our work and the previous work discussed in \autoref{sec: previous work}. Firstly, the authors in \cite{JAWS} propose the bound $\prob{Y\in\intRA{X}} \geq 1-c\alpha(X,\Z)$. This bound, as we discussed earlier, is erroneous for CP techniques that do not provide conditional coverage. Our analysis corrects this bound to $\prob{Y\in\intRA{X}} \geq 1-c\E_{\Z,X}\left[\alpha(X,\Z)\right]$. Secondly, for $m=1$, the approximator $\alpha_\I^m$ given in \autoref{def alphaIm} is a crude high-variance approximation for $\alpha_\I$. Furthermore, this crude approximator is the same as the approximation $\alpha(X,\Z)$ proposed in \cite{JAWS}. Therefore, the estimator in \cite{JAWS} can be viewed as a special case of our estimator, which better captures the randomness of the coverage estimate (defined in \autoref{alphaXZ}) via a hold-out set.
\end{remark}

\begin{algorithm}[tb]
   \caption{Inverse Conformal Prediction (InvCP)}
   \label{alg:ICP}
\begin{algorithmic}
   \STATE {\bfseries Input:} a calibration data set $\Z_0=\{(X_i
, Y_i)\}_{1 \leq i \leq n}$; $\alpha$-calibration set $\Z_0^\alpha = \{X_{i_\alpha}\}_{1 \leq i_\alpha \leq m}$; a pre-defined interval function $\mathcal{I}(X)=\left[a_-(X),a_+(X)\right]$; a weight function $p_i^w(x)$ .
   \FOR {\textbf{each} $X^*\in \Z_0^\alpha$}
   \FOR{$i=1$ {\bfseries to} $n$}
   \STATE Compute $V_{i}^\pm(X^*)$ and $p_i^w(X^*)$; 
   \ENDFOR
   \STATE Compute $\alpha(X^*,\Z_0)$ using \autoref{eq:alpha_XZ_sol};
   \ENDFOR
     \STATE {\bfseries Output:} Return $1-\alpha_\I^m$, where $\alpha_\I^m =\frac{1}{m}\sum_{X^*\in \Z_0^\alpha} \alpha(X^*,\Z_0)$.
\end{algorithmic}
\end{algorithm}

\section{Experimental results}
\label{sec: experiment}
The goal of this section is to empirically test the performance of the proposed risk assessment algorithm, based on different conformal prediction methods for models with varying predictive performance and dataset sizes, under both exchangeable data and covariate shift settings. 

\subsection{Experimental setup}
We simulated a dataset such that the ground truth for the probability of the true label belonging to a specified interval could be estimated with a high degree of accuracy. We constructed the base model using only two covariates $X_1$ and $X_2$ (for simplicity of calculating true probability) with a log-transformation for establishing a non linear relationship. 
The model is defined as -
\begin{equation}
\label{model_eqn}
 Y = X_{1} *  |\log (|X_{2}/100|)| 
    + X_{2} *  |\log (|X_{1}/100|)| 
    + \epsilon   
\end{equation}
where $X_1 \sim N(\mu_1, \sigma_1)$, $X_2 \sim N(\mu_2, \sigma_2)$ and $\epsilon \sim N(0, \sigma_{\epsilon})$. For the experiment, we take the parameters $[\mu_1, \mu_2, \sigma_1, \sigma_2, \epsilon]$ as $[70, 40, 20, 10, 5]$.  An illustration of the relationship between $Y$ and $X_{1}$ and $X_{2}$ in the absence of any noise $\epsilon$ is included in the supplementary material (Figure S.1).

We sampled from the underlying distributions of $X_1, X_2, \epsilon$ to generate a dataset of a fixed size ($n_{total}$). We conducted a series of $B = 100$ trials with different training and test data splits from $n_{total}$ (where $n_{train} = \frac{7}{10} n_{total}, n_{test} = \frac{3}{10} n_{total}$) and fit different models to compute the predicted values on the test set for each trial. For each trial, we then applied the InvCP methods to estimate the probability that $\mathcal{I}(X) = [\hat{\mu}(X) - \tau, \hat{\mu}(X) + \tau]$ over the test set. 

For the given data, the target variable $Y$ has a mean of around $85$ and a standard deviation of around $20$. Based on this, we set $\tau = 10$, i.e. the sensitivity threshold of the absolute difference between the predicted and the true value is set equal to $10$. The supplementary material (Figure S.1) includes an illustrative example of the problem for a random set of 100 test points. For each of the $B$ trials, we then computed the coverage estimates $1-\alpha_\I^m$ following the InvCP algorithm (Supplement S.1) for the \textit{Split conformal\cite{split_cp_1,split_adapt}}, \textit{Jackknife+ \cite{cv_for_cp}}, and \textit{Cross Validation+ (CV+\cite{cv_for_cp})} CP techniques. A \textit{True probability} is also computed as a baseline, which is the empirical probability of the true outcome $Y$ falling inside of the given interval $\mathcal{I}(X)$ over the test sets in all trials.

\subsection{Method effectiveness}
\label{sec: comp CP}
\begin{figure}[hbt!]
\centering
\includegraphics[width = 0.5\textwidth]
{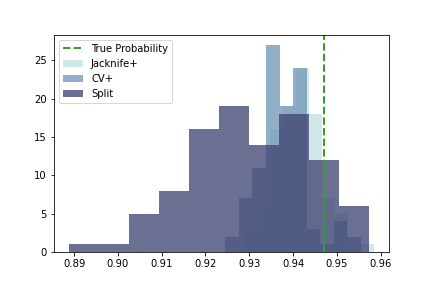}
\caption{\small{Coverage estimates. \textit{Green dashed line denotes averaged empirical coverage of $\mathcal{I}(X)$ over 100 trials (approximated ground-truth probability). The histograms denote the distributions of coverage estimates by the InvCP algorithm based on Jacknife+, CV+, and Split over all trials.}}}
  \label{results1}
\end{figure}

We first run the experiment using a polynomial regression model and a dataset of size 1000. The true coverage over a set of B = 100 trials and corresponding estimates for coverage obtained using the risk assessment method based on different conformal prediction techniques are shown in \autoref{results1}. The plot demonstrates that the average split, jackknife+, and CV+ estimates all lie close to the average true probability estimate, though estimates are generally lower than the true results. This is important to note because this demonstrates that the method works effectively and gives conservative coverage estimates, in line with the methodology of risk assessment. Further, we can observe that the split conformal results have much wider variability as compared to CV+/ jackknife+, which is again in line with what we would expect as split conformal only utilizes half the dataset for training and calibration, thereby making it more variable than CV+/ jackknife+.

\begin{figure}[bt!]
\begin{subfigure}{.495\linewidth}
  \includegraphics[width=\linewidth]{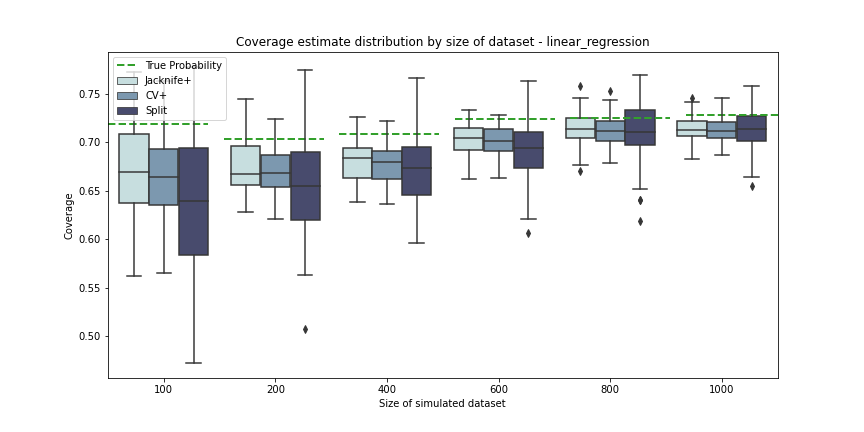}
  \caption{Linear Regression}
  \label{linear_reg_simulated_1000_abs}
\end{subfigure}\hfill 
\begin{subfigure}{.495\linewidth}
  \includegraphics[width=\linewidth]{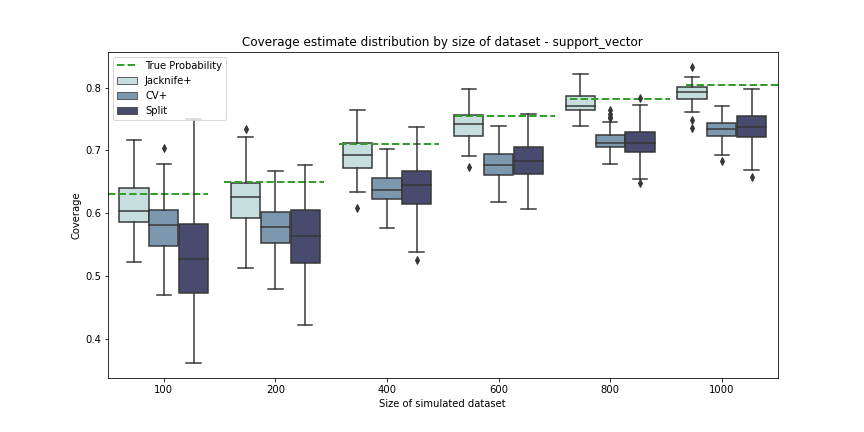}
  \caption{Support Vector Regression (SVR)}
  \label{support_vector_simulated_1000_abs}
\end{subfigure}

\begin{subfigure}{.495\linewidth}
  \includegraphics[width=\linewidth]{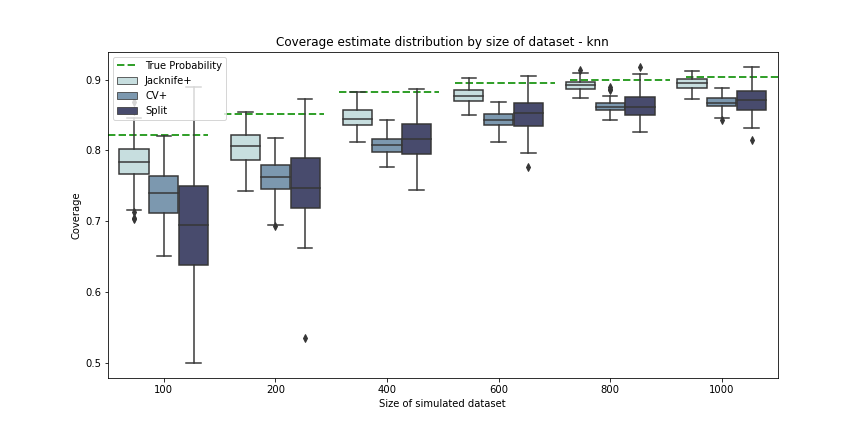}
  \caption{KNN}
  \label{knn_simulated_1000_abs}
\end{subfigure}\hfill 
\begin{subfigure}{.495\linewidth}
  \includegraphics[width=\linewidth]{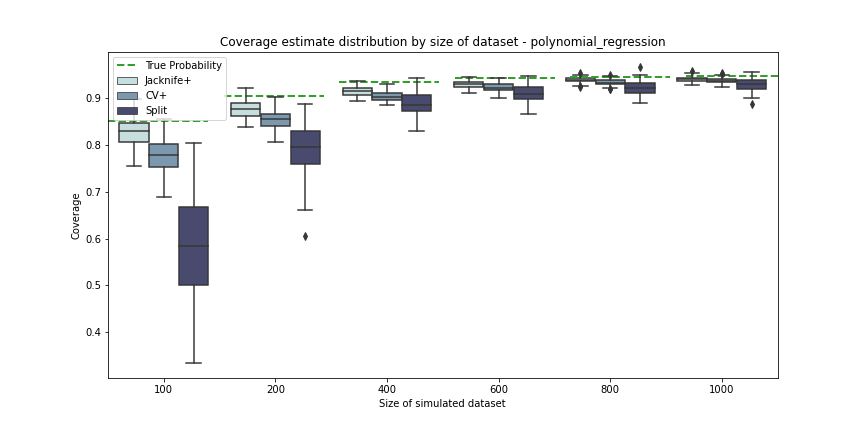}
  \caption{Polynomial Regression}
  \label{polynomial_regression_simulated_1000_abs}
\end{subfigure}

\caption{\small{Comparison by the size of dataset. \textit{Green dashed lines denote the empirical coverages of $\mathcal{I}(X)=[\hat{\mu}(X) - 10, \hat{\mu}(X) + 10]$ where the predictions $\hat{\mu}(X)$ are fitted by different models with different data size, respectively. The boxplots show coverage estimates by the InvCP algorithm based on Jacknife+, CV+, and Split over 100 trials. For any size of data, all methods provide smaller coverage estimates than the ground-truth probability on average.}}}
\label{simulated_size_abs}
\end{figure}

We now proceed to compare how these trends are impacted across models and datasets of different sizes. We consider models of different levels of complexity specifically - linear regression \cite{linear_reg}, support vector regression (SVR) \cite{svr}, K-nearest neighbours (KNN) \cite{knn} with $k = 5$ and polynomial regression \cite{poly_reg}, and data sizes increasing from 100 to 1000. These models and data sizes simulate the various predictive power of $\hat{Y}$, see the predictive performance of these models in the Supplement (Table S.1). The ground true coverage probability of $\mathcal{I}(X)$ is therefore varying accordingly, and we test the accuracy of our risk assessment methods under these scenarios. \autoref{simulated_size_abs} shows the results.

Across all models and dataset sizes, it is evident that all our coverage estimators are lower than the ground-truth probability on average, which proves the conservativeness of our risk assessment methods. In terms of accuracy, Jackknife+ provides closer estimators to the true probability on average in all simulated cases. In terms of variation of estimation, Jackknife+
and CV+ have lower variability (range) over trials than split conformal. We can see that as the size of the dataset increases or the predictiveness of model increases, the width of the coverage estimate spread decrease. This observation makes intuitive sense as a larger quantity of data and more predictive model should reduce variability in estimates. Further, we observe that this improvement is more marked for split conformal compared to jackknife+ and CV+. The latter constitutes an important result as it implies that depending on the size of the dataset, split conformal results can be very close to those from jackknife+ or CV+. 

We compare the performance of the proposed estimators (especially based on weighted and unweighted CP intervals) under similar experimentation but for a dataset with covariate shift. The results show effectiveness of the InvCP algorithm under covariate shift, see the supplementary material S.2.

\section{Conclusions}
We have shown how conformal prediction-based prediction interval techniques can be used to estimate the failure probability of an ML model under both, exchangeability and covariate shift. We theoretically proved that our approach is conservative and experimentally validated its accuracy. Our experiments demonstrate the performance of the risk assessment approach, comparing performance for different model complexities and different sizes of datasets under exchangeability and covariate shift setting (Supplement S.2.2). 

The results obtained reflect the performance of the underlying model, i.e. as model quality improves, the coverage improves, and the estimates obtained are in line with this behaviour.

\appendix

\section{Method}
\subsection{Algorithm details}\label{sec: algorithm}
We provide a walk-through of the proposed InvCP (Inverse Conformal Prediction in \autoref{alg:ICP}). For each element in the $\alpha$-hold-out set $\Z_0^\alpha$, we compute the location and the weights of the point masses appearing in the definition of the prediction interval given in \autoref{interval gen}, i.e. we compute the weights $p_i^w$, and the location of the lower and the upper point masses defined in \autoref{def Vpm}. We then count the number of upper and lower point masses that are below and above, respectively, the $\I(X)$ interval's endpoints. To account for possible co-variate shift, we weigh these point masses, which finally provides us with samples of $\alpha(X,\Z)$. Taking an average over these samples leads to the desired result.

\begin{remark}[Length invariant symmetric intervals]
We consider an interval $\I(X)$ of the form $\I(X) = [\mu(X)-\tau,\mu(X) + \tau]$, where $\tau > 0$. This interval is symmetric around $\mu(X)$, and its length doesn't change with $X$. Furthermore, we assume that the data is exchangeable i.e., there is no co-variate shift and, thus, the weight function $w=1$ in \autoref{weight fun}. For such a case, as shown below, the miscoverage $\alpha(X,Z)$ defined in \autoref{alphaXZ} based on a symmetric prediction interval around $\mu(X)$, e.g., Split-CP, is independent of $X$ for any given $\Z = \Z_0$. Consequently, no $\alpha$-hold-out set is required to collect samples of $\alpha(X,\Z)$, i.e. access to a training and calibration set is sufficient. 

We further elaborate on the above claim. Under exchangeable data, the weights read $p_i^w(X) = \frac{1}{n+1}$ for all $i\in \{1,\dots,n\}$. Applying these weights in \autoref{eq:alpha_XZ_sol}, we find
 \begin{equation}
\alpha(X,\Z_0)=\frac{1}{n+1}\sum_{i=1}^n\mathbbm{1}\{S_i\geq \tau\}= \alpha_0(\Z_0)
 \end{equation}
 where $S$ is the score function $S(x,y) = |y - \mu(x)|$.The sum $\frac{1}{n+1}\sum_{i=1}^n\mathbbm{1}\{S_i\geq \tau\}$ is independent of $X$ and hence, for a given calibration set, so is $\alpha(X,\Z_0)$. 
\end{remark}
 \begin{remark}[Connection to earlier work-continued]
For the particular case discussed above, our estimator $\alpha_\I^m$, for all $m$, would reduce to $\alpha_0(\Z_0)$, which equals $\alpha(X,\Z_0)$. Thus, in this case, our estimator would be the same estimator as that proposed in \cite{JAWS}. We emphasize, however, that the bound $\prob{Y\in\intRA{X}} \geq 1-c\alpha_0(\Z_0)$ resulting from \cite{JAWS} would still be erroneous since it disregards the randomness in the calibration set $\Z$.

 \end{remark}

\section{Experiments}

\subsection{Details of experiments for exchangeable data}

\begin{figure}[hbt!]
\centering
\includegraphics[width = 0.7\linewidth]{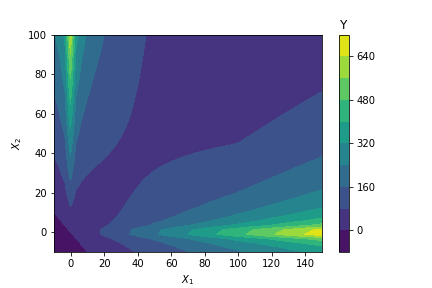}
\caption{\small{Relationship between $X_1$, $X_2$ and $Y$ in the simulated data}}
  \label{simulation_function}
\end{figure}

\begin{figure}[hbt!]
\centering
\includegraphics[width = 0.95\textwidth]
{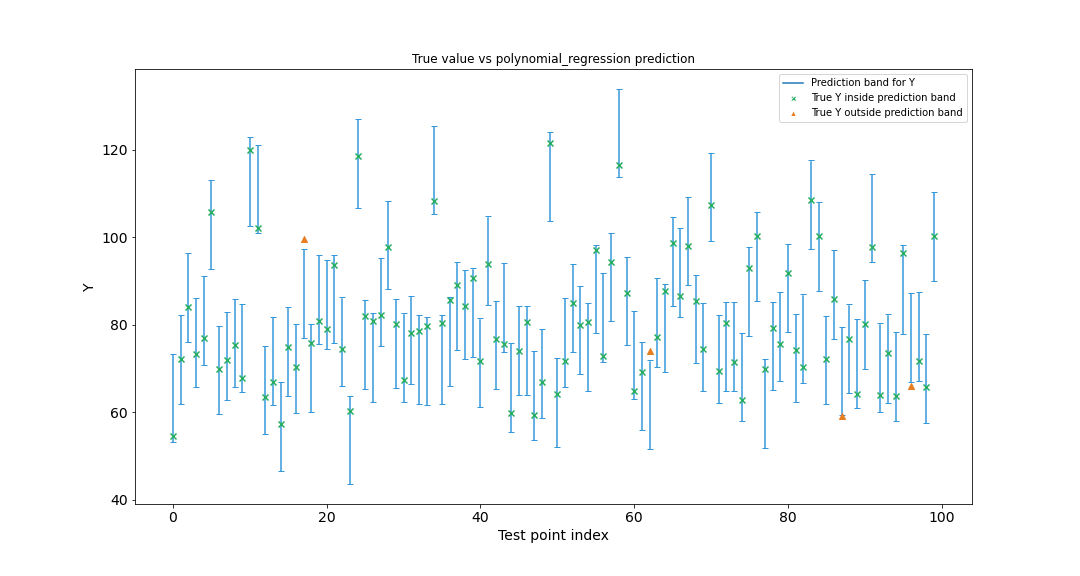}
\caption{\small{Prediction intervals with fixed threshold $\tau$. \textit{The blue intervals denote the pre-defined interval $\mathcal{I}(X_i) = [\hat{\mu}(X_i) - 10, \hat{\mu}(X_i) + 10], i=1, \cdots, 100$. The green cross denote the outcomes of target variable $Y$ that fall inside of $\mathcal{I}(X_i)$, and the red triangles denote the outcomes that fall outside of $\mathcal{I}(X_i)$. This plot shows the empirical marginal coverage of $\mathcal{I}(X)$ is 0.96.}}}
  \label{intervals_fixed_tau}
\end{figure}

This section provides additional details of the experiments in our paper (\autoref{sec: experiment}). \autoref{simulation_function} shows the nonlinear relationship of target variable $Y$ and input variables $X_1$ and $X_2$ in the absence of noise term $\epsilon$. \autoref{intervals_fixed_tau} illustrates the given interval $\mathcal{I}(X)$ and computation of its empirical coverage, which we used as the \textit{true probability} in our experiments for evaluating the performance of proposed estimators. Table \ref{model_comparison_simulated} provides the performance of the predictive models used in the experiments.

\begin{center}
\begin{tabular}{|l|lll|}
\hline
Model & RMSE & MAE & R2 \\
\hline
Linear Regression & 10.53 & 7.46 & 0.46 \\
SVR & 9.23 & 6.49 & 0.62 \\
KNN & 6.46 & 5.05 & 0.82 \\
Polynomial Regression & 5.37 & 4.42 & 0.89 \\
\hline
\end{tabular}
\captionof{table}{\small{Simulated data - Model performance metrics. \textit{As model complexity increases from Linear Regression to Polynomial Regression, the model performance improves, as observed from the metrics in the table.}}}
    \label{model_comparison_simulated}
\end{center}

\subsection{Experiments under covariate shift}

\begin{figure}[hbt!]
\includegraphics[width=\linewidth]{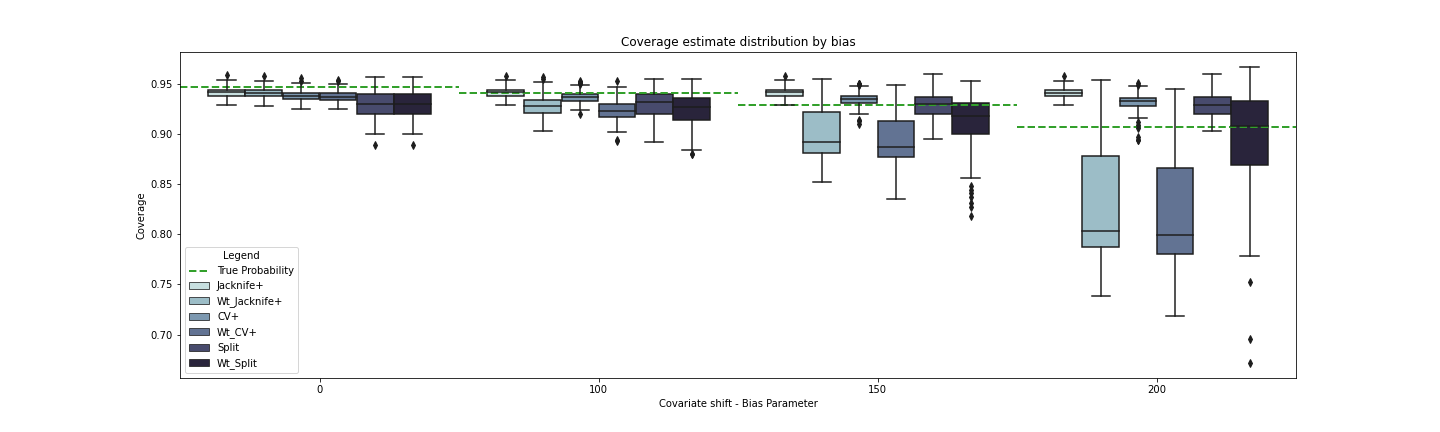}
\caption{\small{Coverage estimates for varying strength of bias parameter.  \textit{Green dashed lines denote the empirical average coverage of $\mathcal{I}(X)=[\hat{\mu}(X) - 10, \hat{\mu}(X) + 10]$ where the predictions $\hat{\mu}(X)$ is fitted by Polynomial Regression. The boxplots show coverage estimates by the InvCP algorithm based on different CP methods over 100 trials. As the bias parameter in simulated covariate shift increases, only weighted CP methods can provide conservative/smaller coverage estimates comparing with ground-truth probability on average.}}}
  \label{coverage_by_bias}
\end{figure}

We further evaluate the proposed risk assessment under the covariate shift setting in \autoref{sec: cov shift}. We simulate an \textit{exponential tilting} between $X_{\text{train}}$ and $X_{\text{test}}$; similar covariate shifts are also considered in previous work \cite{JAWS,CP_covariate}. Let $\tilde{w}(x)=\exp(-\beta/100 \log(X_1)+\beta/100\log(X_2))$, where $\beta$ is a parameter to control the scale of covariate shift (e.g., $\beta=0$ indicates no shift). Then, we re-sample the original testing points with the probabilities $\tilde{w}(x)/||\tilde{w}(x)||_1$ to get the shifted data $\tilde{X}_{\text{test}}$. Hence, the likelihood ratio of the covariate distributions, $w(x)=d\tilde{P}_X(x)/dP_X(x)=\tilde{w}(x)/||\tilde{w}(x)||_1$. We increase the shift parameter $\beta$ from 0 to 200 to simulate the shift from mild to strong. We use a dataset of $n_{\text{total}}=1000$ and a polynomial predictive model. We assume the likelihood ratio of co variate distributions $w$ is known for simplicity. In practice, a model can be fitted to estimate this likelihood ratio, and we refer the details to \cite{CP_covariate}. Apart from the methods in previous sections, we also applied the weighted CP intervals (\textit{Weighted Jacknife+}\cite{JAWS}, \textit{Weighted CV+}\cite{JAWS}, \textit{Weighted Split}\cite{CP_covariate}) in InvCP algorithm, and evaluate the performance of the generated coverage estimates. The results are shown in \autoref{coverage_by_bias}.

All methods can give conservative and accurate estimates when the shift degree is small, although the split conformal-based methods have more variability than others. As the covariate shift degree increases, only the weighted methods (weighted Jacknife+/CV+/Split) can give conservative estimates. In contrast, all unweighted methods over-estimate the coverages on average (i.e., under-estimate the risks of true outcomes falling outside of given intervals). This shows that the weighted methods can adjust to the underlying covariate shift better than their unweighted competitors. One may also notice that the weighted methods have large variances when there is a significant covariate shift. This can be explained by the large variance of the shifted testing sets created by data-dependent weights: for each train/test split, the weights are calculated based on the test set, and a shifted test set is obtained by resampling with these simulated weights. As the weighted methods adjust to the empirical testing covariate distribution closely, the variances of coverage estimates increase proportionally to the variance of the empirical coverage probabilities as the shift parameter $\beta$ increases, which is another validation of the accuracy of the weighted methods under covariate shift.

\subsection{Space variant $\tau$}
\begin{figure}[hbt!]
\includegraphics[width=0.95\linewidth]{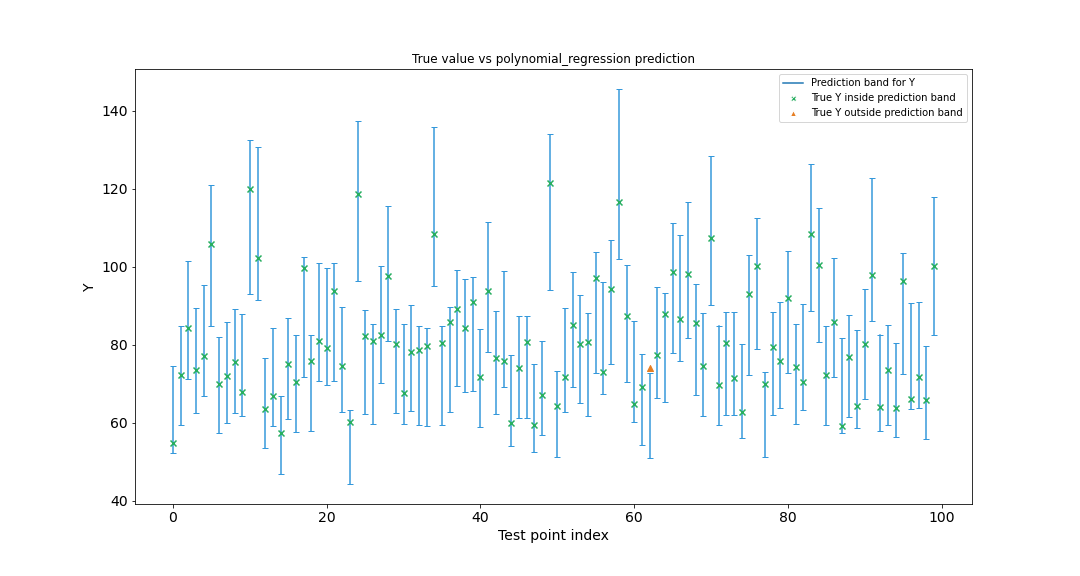}
\caption{\small{Prediction intervals - varying threshold $\tau$}}
  \label{intervals_varying_tau}
\end{figure}

\begin{figure}[hbt!]
\centering
\includegraphics[width = 0.85\linewidth]{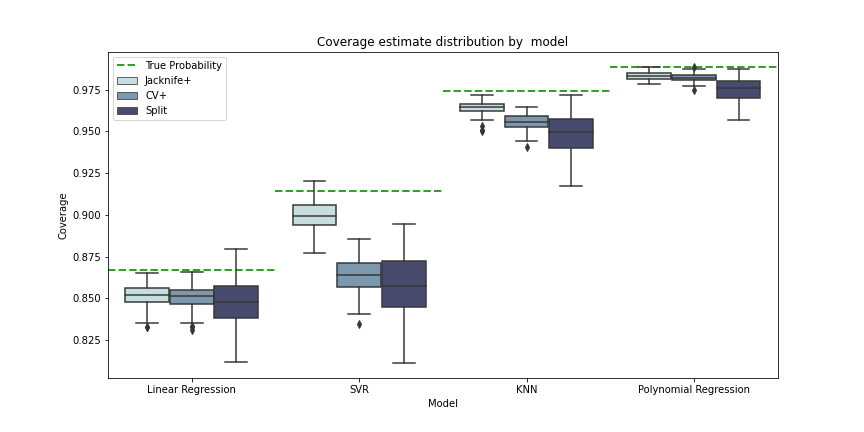}
\caption{\small{Comparison by model for variant $\tau$}}
  \label{result: variant1}
\end{figure}

\begin{figure}[bt!]
\begin{subfigure}{.495\linewidth}
  \includegraphics[width=\linewidth]{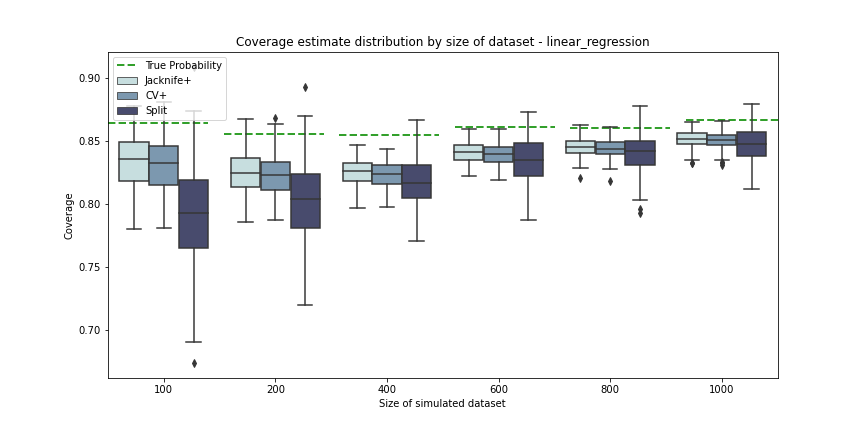}
  \caption{Linear regression}
  \label{linear_reg_simulated_1000_abs}
\end{subfigure}\hfill 
\begin{subfigure}{.495\linewidth}
  \includegraphics[width=\linewidth]{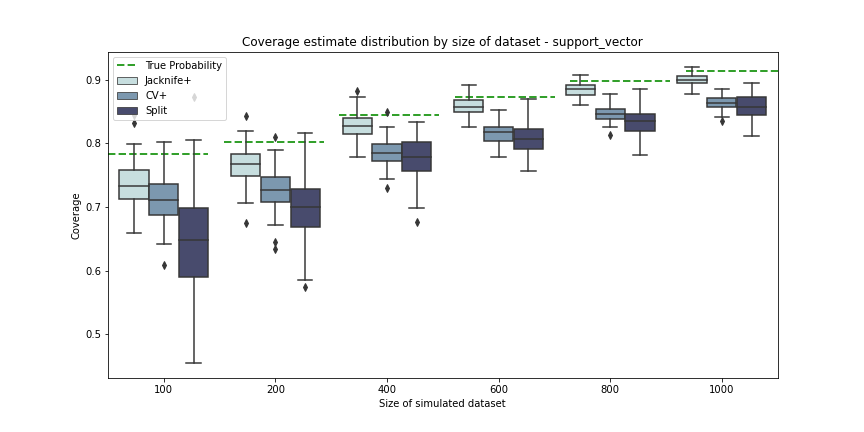}
  \caption{Support vector}
  \label{support_vector_simulated_1000_abs}
\end{subfigure}

\begin{subfigure}{.495\linewidth}
  \includegraphics[width=\linewidth]{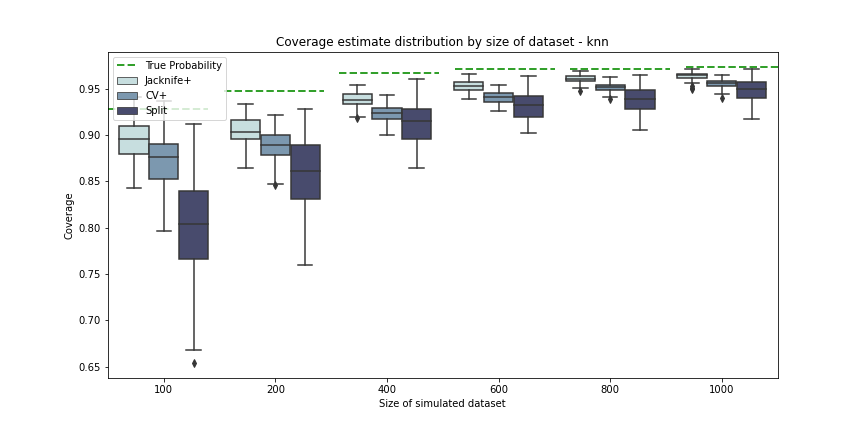}
  \caption{KNN}
  \label{knn_simulated_1000_abs}
\end{subfigure}\hfill 
\begin{subfigure}{.495\linewidth}
  \includegraphics[width=\linewidth]{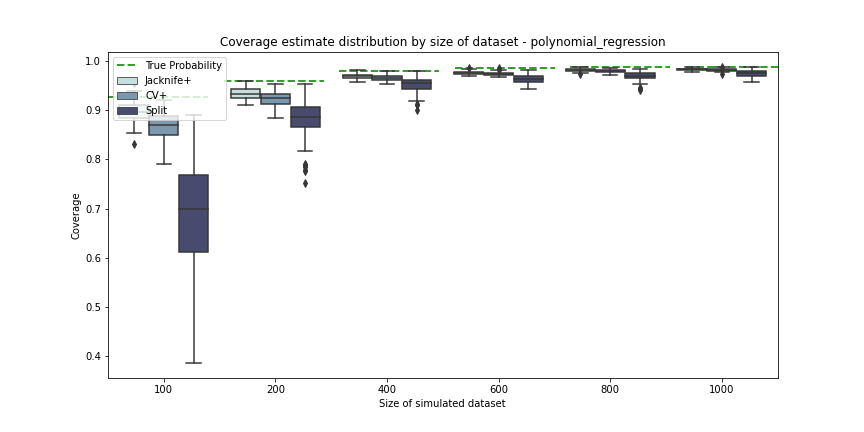}
  \caption{Polynomial regression}
  \label{polynomial_regression_simulated_1000_abs}
\end{subfigure}
\caption{\small{Comparison by size of data for variant $\tau$}}
\label{result: variant2}
\end{figure}

We have considered the probability $Y \in [\hat{\mu}(X) - \tau, \hat{\mu}(X) + \tau ]$. Here, $\tau$ is fixed and doesn't change with $Y$. However, in many real life applications, we may wish to consider the problem setup where the user defined threshold is not space invariant but may be a function of $Y$. For example, in many industry and engineering applications, error thresholds are expressed as a percentage of $Y$ i.e the model prediction $\hat{\mu}$ is considered correct if the relative difference between the prediction and the true value does not exceed the defined threshold $\tau$ i.e $ | \frac{Y - \hat{\mu}}{\hat{\mu}} | \leq \tau$ which implies $Y \in [\hat{\mu}  - |\hat{\mu}| \tau , \hat{\mu} + |\hat{\mu}| \tau) ]$. We would like to note here that the above method can be easily extended to this problem with the modification that $\tau$ will now be different for each test data point and correspond to $ \tau^{'} = |\hat{\mu}| \tau$. \autoref{intervals_varying_tau}  shows the adaptive prediction bands for $Y_{test}$ for the simulated dataset with $\tau = 0.175$. As we can see, the size of the prediction bands varies with the predicted value. The performance are similar to the fixed $\tau$ case, and we provide the results for references in \autoref{result: variant1}-\autoref{result: variant2}.

\section{Discussions and future work}
The current paper proposes an interval-generation based method for risk assessment of regression models. We restrict this work to JAW and weighted split conformal prediction intervals as they provide theoretical coverages under covariate shift. Other conformal prediction intervals such as Conformalized Quantile Regression \cite{cqr} or Conformalizing Bayes \cite{conformal_pred} can also be applied to our framework following the proposed InvCP algorithm. However, their theoretical properties under covariate shift are unclear and need further explorations. Furthermore, the recent development of adaptive conformal inference under arbitrary distribution shifts \cite{arbitrary} provides the potential of conducting risk assessment in an online fashion. Deriving conservative risk assessment methods under arbitrary shifts is important as many ML models are used in the fast-changing areas such as finance and economics, where the market and customers behaviours can shift abruptly. In the future, we aim to extend the InvCP algorithm to improve the adaptivity of our method in these more challenging data scenarios.

\medskip
\small
\bibliography{example_paper}

\begin{thebibliography}{10}

\bibitem{battery_1}
Man-Fai Ng, Jin Zhao, Qingyu Yan, Gareth~J Conduit, and Zhi~Wei Seh.
\newblock Predicting the state of charge and health of batteries using data-driven machine learning.
\newblock {\em Nature Machine Intelligence}, 2(3):161--170, 2020.

\bibitem{battery_2}
Darius Roman, Saurabh Saxena, Valentin Robu, Michael Pecht, and David Flynn.
\newblock Machine learning pipeline for battery state-of-health estimation.
\newblock {\em Nature Machine Intelligence}, 3(5):447--456, 2021.

\bibitem{radiationML}
Elizabeth Huynh, Ahmed Hosny, Christian Guthier, Danielle~S Bitterman, Steven~F Petit, Daphne~A Haas-Kogan, Benjamin Kann, Hugo~JWL Aerts, and Raymond~H Mak.
\newblock Artificial intelligence in radiation oncology.
\newblock {\em Nature Reviews Clinical Oncology}, 17(12):771--781, 2020.

\bibitem{conditionalCP}
Vladimir Vovk.
\newblock Conditional validity of inductive conformal predictors.
\newblock In {\em Asian conference on machine learning}, pages 475--490. PMLR, 2012.

\bibitem{conformal_pred_intro}
Vladimir Vovk, Alexander Gammerman, and Glenn Shafer.
\newblock {\em Algorithmic learning in a random world}, volume~29.
\newblock Springer, 2005.

\bibitem{conformal_pred}
Anastasios~N Angelopoulos and Stephen Bates.
\newblock A gentle introduction to conformal prediction and distribution-free uncertainty quantification.
\newblock {\em arXiv preprint arXiv:2107.07511}, 2021.

\bibitem{cp_res_score}
Jing Lei, Max G’Sell, Alessandro Rinaldo, Ryan~J Tibshirani, and Larry Wasserman.
\newblock Distribution-free predictive inference for regression.
\newblock {\em Journal of the American Statistical Association}, 113(523):1094--1111, 2018.

\bibitem{cqr}
Yaniv Romano, Evan Patterson, and Emmanuel Candes.
\newblock Conformalized quantile regression.
\newblock {\em Advances in neural information processing systems}, 32, 2019.

\bibitem{score_function_1}
Danijel Kivaranovic, Kory~D Johnson, and Hannes Leeb.
\newblock Adaptive, distribution-free prediction intervals for deep networks.
\newblock {\em International Conference on Artificial Intelligence and Statistics}, pages 4346--4356, 2020.

\bibitem{score_function_2}
Matteo Sesia and Emmanuel~J Cand{\`e}s.
\newblock A comparison of some conformal quantile regression methods.
\newblock {\em Stat}, 9(1):e261, 2020.

\bibitem{score_function_3}
Victor Chernozhukov, Kaspar W{\"u}thrich, and Yinchu Zhu.
\newblock Distributional conformal prediction.
\newblock {\em Proceedings of the National Academy of Sciences}, 118(48):e2107794118, 2021.

\bibitem{split_cp_1}
Harris Papadopoulos, Kostas Proedrou, Volodya Vovk, and Alex Gammerman.
\newblock Inductive confidence machines for regression.
\newblock {\em Lecture notes in computer science}, pages 345--356, 2002.

\bibitem{split_adapt}
Jing Lei, Alessandro Rinaldo, and Larry Wasserman.
\newblock A conformal prediction approach to explore functional data.
\newblock {\em Annals of Mathematics and Artificial Intelligence}, 74:29--43, 2015.

\bibitem{classification_avg_size}
Mauricio Sadinle, Jing Lei, and Larry Wasserman.
\newblock Least ambiguous set-valued classifiers with bounded error levels.
\newblock {\em Journal of the American Statistical Association}, 114(525):223--234, 2019.

\bibitem{cp_classficiation_image}
Anastasios Angelopoulos, Stephen Bates, Jitendra Malik, and Michael~I Jordan.
\newblock Uncertainty sets for image classifiers using conformal prediction.
\newblock {\em arXiv preprint arXiv:2009.14193}, 2020.

\bibitem{old_tutorial}
Glenn Shafer and Vladimir Vovk.
\newblock A tutorial on conformal prediction.
\newblock {\em Journal of Machine Learning Research}, 9(3), 2008.

\bibitem{CP_covariate}
Ryan~J Tibshirani, Rina Foygel~Barber, Emmanuel Candes, and Aaditya Ramdas.
\newblock Conformal prediction under covariate shift.
\newblock {\em Advances in neural information processing systems}, 32, 2019.

\bibitem{beyond_exchangebility}
Rina~Foygel Barber, Emmanuel~J Candes, Aaditya Ramdas, and Ryan~J Tibshirani.
\newblock Conformal prediction beyond exchangeability.
\newblock {\em arXiv preprint arXiv:2202.13415}, 2022.

\bibitem{JAWS}
Drew Prinster, Anqi Liu, and Suchi Saria.
\newblock Jaws: Auditing predictive uncertainty under covariate shift.
\newblock {\em Advances in Neural Information Processing Systems}, 35:35907--35920, 2022.

\bibitem{Goodfellow-book}
Ian Goodfellow, Yoshua Bengio, and Aaron Courville.
\newblock {\em Deep learning}.
\newblock MIT press, 2016.

\bibitem{covariateML}
Hidetoshi Shimodaira.
\newblock Improving predictive inference under covariate shift by weighting the log-likelihood function.
\newblock {\em Journal of statistical planning and inference}, 90(2):227--244, 2000.

\bibitem{covariate_shift_1}
Masashi Sugiyama, Matthias Krauledat, and Klaus-Robert M{\"u}ller.
\newblock Covariate shift adaptation by importance weighted cross validation.
\newblock {\em Journal of Machine Learning Research}, 8(5), 2007.

\bibitem{data_shift_1}
Joaquin Quinonero-Candela, Masashi Sugiyama, Anton Schwaighofer, and Neil~D Lawrence.
\newblock {\em Dataset shift in machine learning}.
\newblock Mit Press, 2008.

\bibitem{data_shift_2}
Yaniv Ovadia, Emily Fertig, Jie Ren, Zachary Nado, David Sculley, Sebastian Nowozin, Joshua Dillon, Balaji Lakshminarayanan, and Jasper Snoek.
\newblock Can you trust your model's uncertainty? evaluating predictive uncertainty under dataset shift.
\newblock {\em Advances in neural information processing systems}, 32, 2019.

\bibitem{data_shift_3}
Dennis Ulmer, Lotta Meijerink, and Giovanni Cin{\`a}.
\newblock Trust issues: Uncertainty estimation does not enable reliable ood detection on medical tabular data.
\newblock {\em Machine Learning for Health}, pages 341--354, 2020.

\bibitem{cv_for_cp}
Rina~Foygel Barber, Emmanuel~J. Candes, Aaditya Ramdas, and Ryan~J. Tibshirani.
\newblock Predictive inference with the jackknife+.
\newblock {\em The Annals of Statistics}, 49:486--507, 2021.

\bibitem{linear_reg}
Xiaogang Su, Xin Yan, and Chih-Ling Tsai.
\newblock Linear regression.
\newblock {\em Wiley Interdisciplinary Reviews: Computational Statistics}, 4(3):275--294, 2012.

\bibitem{svr}
Debasish Basak, Srimanta Pal, and Dipak Patranabis.
\newblock Support vector regression.
\newblock {\em Neural Information Processing – Letters and Reviews}, 11, 10 2007.

\bibitem{knn}
Shreya Kohli, Gracia~Tabitha Godwin, and Siddhaling Urolagin.
\newblock Sales prediction using linear and knn regression.
\newblock {\em Advances in Machine Learning and Computational Intelligence: Proceedings of ICMLCI 2019}, pages 321--329, 2021.

\bibitem{poly_reg}
Eva Ostertagov{\'a}.
\newblock Modelling using polynomial regression.
\newblock {\em Procedia Engineering}, 48:500--506, 2012.

\bibitem{arbitrary}
Isaac Gibbs and Emmanuel Candes.
\newblock Adaptive conformal inference under distribution shift.
\newblock volume~34, pages 1660--1672, 2021.

\end{thebibliography}
\bibliographystyle{unsrt}

\end{document}